\documentclass[conference]{IEEEtran}
\IEEEoverridecommandlockouts
\usepackage{cite}
\usepackage{amsmath,amssymb,amsfonts}
\usepackage{algorithmic}
\usepackage{graphicx}
\usepackage{textcomp}
\usepackage{xcolor}
\def\BibTeX{{\rm B\kern-.05em{\sc i\kern-.025em b}\kern-.08em
    T\kern-.1667em\lower.7ex\hbox{E}\kern-.125emX}}
\begin{document}

\title{Abstractive Information Extraction from  Scanned Invoices (AIESI) using End-to-end Sequential Approach\\
}

\author{\IEEEauthorblockN{Shreeshiv Patel}
\IEEEauthorblockA{\textit{Bachelor of Technology in Information Technology} \\
\textit{Institute of Technology, Nirma University}\\
Ahmedabad, India \\
16bit101@nirmauni.ac.in}
\and
\IEEEauthorblockN{Dvijesh Bhatt}
\IEEEauthorblockA{\textit{Department of Computer Science and Engineering} \\
\textit{Institute of Technology, Nirma University}\\
Ahmedabad, India \\
dvijesh.bhatt@nirmauni.ac.in}
\and
}

\maketitle

\begin{abstract}
Recent proliferation in the field of Machine Learning and Deep Learning allows us to generate OCR models with higher accuracy. Optical Character Recognition(OCR) is the process of extracting text from documents and scanned images. For document data streamlining, we are interested in data like, Payee name, total amount, address, and etc. Extracted information helps to get complete insight of data, which can be helpful for fast document searching, efficient indexing in databases, data analytics, and etc. Using AIESI we can eliminate human effort for key parameters extraction from scanned documents. Abstract Information Extraction from Scanned Invoices (AIESI) is a process of extracting information like, date, total amount, payee name, and etc from scanned receipts. In this paper we proposed an improved method to ensemble all visual and textual features from invoices to extract key invoice parameters using Word wise BiLSTM.

\end{abstract}

\begin{IEEEkeywords}
OCR, AIESI\end{IEEEkeywords}

\section{Introduction}
OCR( Optical character recognition) is the process of identifying text characters from scanned documents.\cite{tesseract}\cite{tesseract1} Scanned documents can be Invoice, bill, and receipts. Scanned OCR identifies characters from scanned documents which can be structured or semi-structured. Document can be in multi format, it may be .pdf, .jpg, or any image format. For powering, designing or implementing data driven machine learning models or operations, extracting text from OCR is not enough. We have to extract information such as, Date, Payee name, Total amount, Product list, and etc. Extracting primary parameters from documents can play an important role in many services and applications, like digitizing documents, converting documents information into structured or semi-structured databases. This database can help for archiving, rapid indexing, comparative analysis of data. \cite{icdar2019}. AIESI(Abstract Information Extraction from Scanned Invoices) plays an important role where we have to deal with document intensive tasks in accounting, financial, law firms and medical documents. AIESI helps in automation of data archiving and streamlining primary parameters from documents. We have seen lots of breakthrough research in the domain of OCR with respect to model accuracy and latency. For projection of AIESI in commercial application, we need to have high accuracy and low latency for processing large numbers of documents. In the current ecosystem, data extraction from documents is done by manual processing. Manual processing can be biased and ambiguous in identifying key parameters. AIESI can solve the above problems. It has higher accuracy than human and less ambiguity in identifying key parameters. Manual work has some time overload, while millions of documents can be processed in hours with higher accuracy. When we have to deal with multilingual documents, human understanding of multi-language has some limitations but AIESI design models that can support multilingual documents.  

 OCR extracts text from invoice, but for extracting key parameters this (extracted text from invoice (ETFI) is not enough.  AIESI task is challenging since, model cannot generalise the structure of the document, we may have a heterogeneous structure of invoices.  For aforementioned task, we proposed a model that extract different richer features from invoices and ensemble to find key parameters. Most methods for KIPE use textual information from a detected bounding box using character level sequence tagging \cite{ref14}, inspired from Name Entity Recognition(NER) architecture.\cite{nameEntity} We can get complete insight and extract key invoice parameters correctly by adding spatial and visual features for KIPE.  Our end-to-end sequential model can solve aforementioned problems. We can use our model for any unknown structure of documents.

\section{Related Work}
With recent proliferation, detecting text from scanned documents or images have become accurate and robust. Many frameworks and API developed for aforementioned tasks, (.i.e text detection)Tesseract OCR engine\cite{tesseract}\cite{tesseract1}, Amazon Textract, Google vision API, and many other APIs and frameworks. Extracting text with its spatial information is now become a mature. However extracting key parameter from document with good accuracy and with less latency, still an challenge for services and application where high accuracy is required.\cite{icdar2019}

\subsection{Pattern Matching}

\begin{figure}
    \centering
     \includegraphics[height=1.3cm]{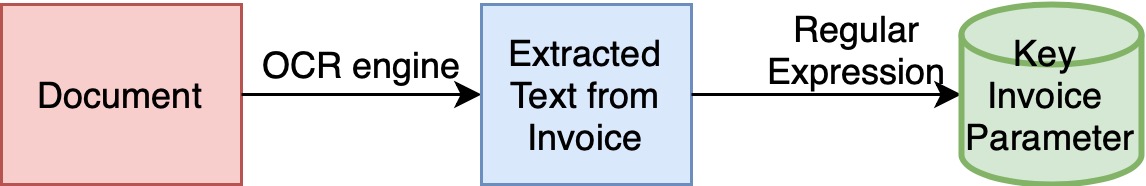}
    \caption{Architecture of rule based KIPE from OCR}
    \label{rule_based}
\end{figure}

Brute force approach for Key Invoice Parameter Extraction(KIPE) is to find required text patterns from Extracted Text From Invoice(ETFI). Figure \ref{rule_based} depicts rule based KIPE architecture. Such as for finding a parameter date, we will look for a NUM/NUM/NUM or NUM/TEXT/NUM in extracted text. For getting key parameters we iteratively search for required patterns from ETFI.  We can also achieve our goal by iteratively searching a KEY in ETFI, and when we hit the key then the surrounding text might be our required parameter corresponding to the key. Example, generally price is preceded or preceded by “TOTAL” keyword or “AMOUNT”. Then we will find a pattern for this key in ETFI then we can search for corresponding parameters surrounding the key's spatial location. 

Using rule based pattern matching KIPE is easy and fast but with very less accuracy. We cannot generalise document structure. For industries where higher standards of accuracy are required this approach can not work. Invoices can have multiple instances of text pattern corresponding to key, in such cases this method can lead to ambiguity.

\subsection{Classification approach for KIPE}
We have word extracted from ETFI. Using classification algorithms, like SVM, Decision tree, logistic regression we can classify each bounding box into class, like address, company, date, total, or none. Since we have multiple classes, our task can be solved by to multi class classifier. We vectorize each bounding box by word embedding using pre-trained language model, followed by classification approach. Here we are assuming each bounding- box is atomic in nature, i.e. each bounding box corresponds to one class only. Classification model classify each bounding box into different classes. There can be multiple labels with price, like {tax:101.00}, {price:111,00}, {discount:11.00}. It can lead to ambiguity in finding required key-value pair as we have multiple bounding box with same key tag. This approach cannot give satisfactory result as we are not using any sequential features. We know that price tag, generally preceded or succeeded by "TOTAL" or "AMOUNT" key-word. 

\subsection{End2end sequential model for KIPE}
We are facing major challenges in KIPE for heterogeneous structure of Invoices. Rule based brute force technique unable to generalise. Recent proliferation in deep learning solves the above mentioned problems of heterogeneous structure of invoices. Using en2end sequential model we can learn from the previous sequence and classify current word into different key-values pairs with higher accuracy. We can apply this knowledge at character level and word level. 

\section{ Method }
In this paper, we introduce an improved method for KIE from extracted text from invoices. We ensemble different features from invoices and followed by bouding box level multi-class classification by BiLSTM. We are exploiting visual, textual and spatial information from invoices. We vectorize each text block with semantic features from text block, spatial features from TB, and visual features from TB. Concatenate each vector, features, then train different classification models for each TB. Each TB is classified into different classes, for our use-cases for invoices, we can classify each of them into classes, Date, Address, Price, Company name, and None. None represent no class. Each classification is followed by a softmax layer for finding probability of each class. Here we are taking one underlying assumption that each TB is atomic in nature, i.e. each TB belongs to only one class, this can be no followed when we have invoice/bills with highly dense information, where there can be possibility that ETFO extract text from different classes into one block.

\subsection{Textual features}

ETFI gives coordinates of bounding box and text within text. Our task is to classify each bounding box into different classes, like Address, company, date, total, invoice number, and etc. 
    After ETFI, we can use semantic features from extracted text. Many model proposed for learning vector space representations of words in capturing fine-grained semantic and syntactic regularities using vector arithmetic\cite{glove}. For embedding text into vectors we can use any pre-trained static word embedding model, like word2vec, GloVe, FastText, and etc. We vectorize each text  with word-embedding. If a bounding-box contains multiple words, then the vector representation of the bounding box can be found by taking a simple average of each word. Bounding box $B_{ij}$ (jth bounding box from ith invoices) may have multiple words, $W_1, W_2,, W_k,$ then we can represent each word by vectors as $v_1, v_2, v_3 ,v_k$ using a pre-trained language model. For representing each bounding box, we are taking simple average.
    Textual features of bounding box can be shown as, $TF_{ij}$(Textual feature of jth bounding box of ith invoice)
    \[TF_{ij}=\sum_{n=1}^{k} v_i /K\]
    
      Taking a simple average, we are giving equal weights to all words, but some of words have more importance to current context semantically and syntactically. Using weighted average, we can represent bounding box with more precisely \cite{sanjiv}.
    
Alternative to above approach, we can do bouding box embedding using dynamic word embedding using pre-trained language model, like BeRT, ROBERT, XLNeT, and etc.  Thus using transformer or pre-trained model that helps to represent semantic-textual information of each bounding box into vectors.

\subsection{Spatial information}
\begin{figure}
    \centering
     \includegraphics[height=8cm]{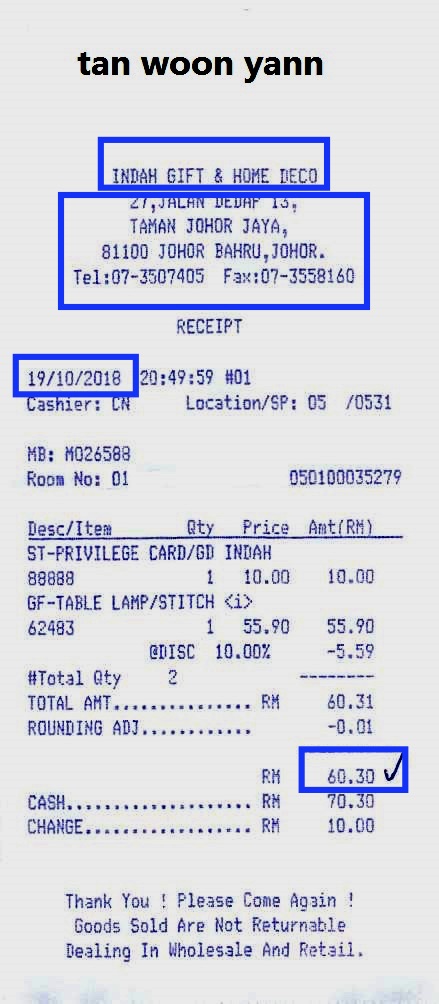}
    \caption{Spatial features from invoices}
    \label{spatial}
\end{figure}

Many traditional approaches are not using spatial information for KIPE. 
In Figure \ref{spatial} we can find some deferential patterns in invoices and bills. Generally company name is one the top of invoice and bills with relatively bold size. Date, invoice number is relatively in the upper part of invoices and resume. Total amount is generally at the bottom-right of invoice and bills. Thus we can hand-pick various differential features from Invoices and can leverage its usage for classifying each bounding box into different classes. We find spatial features, i.e. relative position in invoice and bills. In this approach, we have hand-picked features from the data-set. Each bounding box is annotated with enclosing coordinates and text. For richer spatial features, we can use graph convolutional networks \cite{gcn}\cite{graphConv}. Graph convolutional networks help to extract relative spatial features respect to other bouding box. Graph Convolutional Neural Networks (graph CNNs) have been widely used for graph data representation and semi-supervised learning tasks \cite{11pick}\cite{graphConv1}. However for our proposed method we are using hand-picked but important features from invoices for KIPE. 

Spatial Features used for our method
\begin{itemize}
    \item Bounding box coordinates. Each bounding box is represented by 4 points. Coordinates of bounding box(Bij)  can be is represented as \[C_{ij}={(x_{1ij},y_{1ij}), (x_{2ij},y_{2ij}),(x_{3ij},y_{3ij}), (x_{4ij},y_{4ij})}\]
    
    \item Area of bounding box. Enclosed areas of address and company name bounding box are relatively larger than that of date, invoice number, and price. Area of the bounding box can play a significant role for bounding box classification problems. Area of bounding box calculated as,
    \[A_{ij}=abs(x_{1ij} - x_{2ij}) * abs(y_{1ij} - y_{3ij}) \]
    \item Number of character per unit area. Font size of the company name is relatively larger than other keys. We can define one more feature as the number of characters per unit area of the bounding box, i.e density of character in the bounding box. Let the number of characters in bounding box is countij and ares be $A_{ij}$, previously calculated then,
    \[ D_{ij} = count_{ij}/ A_{ij}\]

\end{itemize}

\subsection{Visual features}
\begin{figure}[h]
    \centering
     \includegraphics[height=6cm]{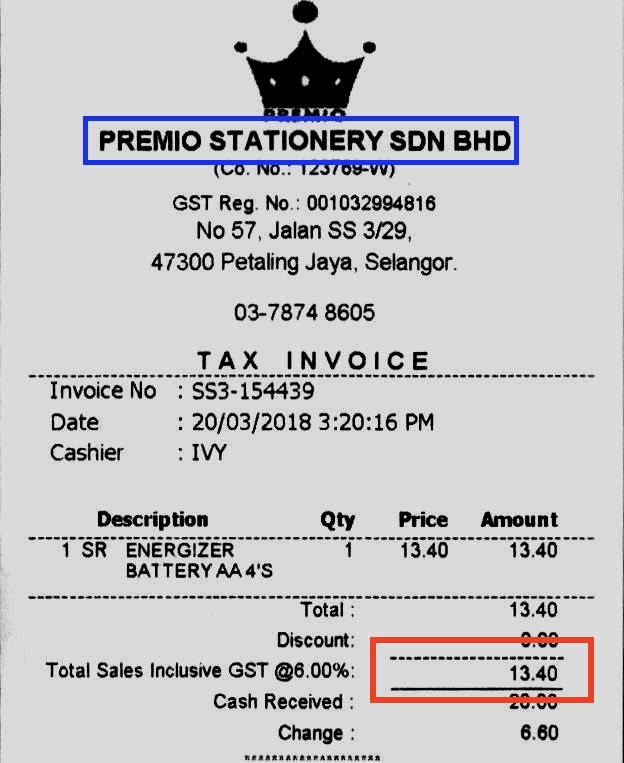}
    \caption{Visual features for a bounding box}
    \label{visual}
\end{figure}
For richer representation it is crucial to extract visual information too. When we deal with  visual features, we mean to how characters are oriented within text-block, color density, color of text, background color of text-block, and etc. These properties play key role for identifying key-value pair in invoices/bills. Using pre-trained image model like, alexnet, resnet, vgg we can extract visual features from text-block. In Figure \ref{visual}, in which color density of company name is relatively higher compared to other bounding boxes. Visual features play a vital role when we invoice with different color depth and with text with different style and density. We used an end-to-end, multimodal, fully convolutional network for extracting semantic structures from document images. We consider document semantic structure extraction as a pixel-wise segmentation task, and propose a unified model that classifies pixels based not only on their visual appearance, as in the traditional page segmentation task, but also on the content of underlying text \cite{featureCNN}.

 Most KIPE architectures are not exploiting visual features for extraction purposes.  In our proposed method, we are assembling visual features for KIPE. Each bounding box has its differential visual features from others, these features can be used to discriminate different classes for KIPE. Convolutional neural network can be used for extracting visual features from bounding box. Common problems we are facing for extracting visual features from pre-trained CNN model are, our bounding box dimens are considerably small than that of pre-trained model originally trained. We need to resize the bounding box according to pre-trained model input-size, which might become an obstacle to get accurate and precise features from the bounding box. For our proposed method, since we have limited number of data points we tried to use transfer learning on different CNN model like, VGG19, ResNet, and inception. Followed by fine tune with it. Output layer of size 5 for our data-set. Since we are interested to extract rich visual features from bounding box, after successful training, we removed trailing layers. Thus, using CNN like architecture we can get image embedding for bounding box.

\subsection{Sequential features}

\begin{figure}[h]
    \centering
     \includegraphics[height=6cm]{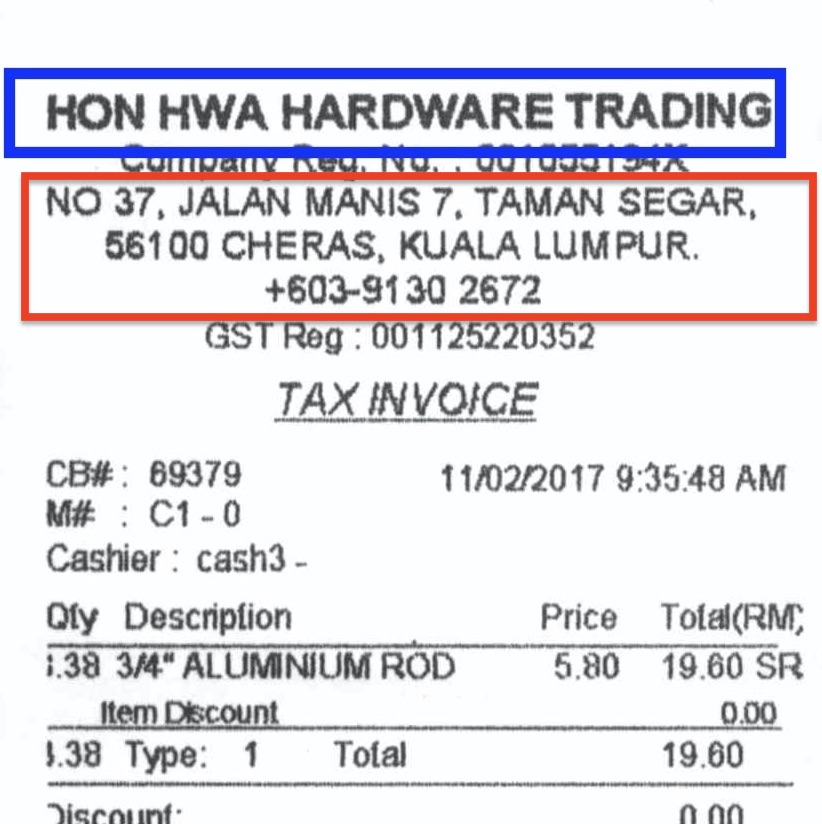}
    \caption{Sequential features in invoices}
    \label{sequential}
\end{figure}
Sequential features consist of relative sequence of text block. Like generally total amount is preceding or succeeded by word like "TOTAL" or "PRICE. Using sequence tagging architecture like LSTM of Recuurrent neural network we can exploiting relative order of text, i.e. sequential features of text for finding KIPE. In Figure \ref{sequential}, depicts generally address name is succeeded by company name. This is sequential information of bounding box text can be exploit using end2end sequential tagging model. In proposed method, we are using bi-directional LSTM followed by CRF for tagging each buding box into different classes. LSTM.\cite{crf}\cite{bilstm-crf}\cite{end2endseq}

\begin{figure*}[h]
    \centering
     \includegraphics[width=\textwidth, height=5cm]{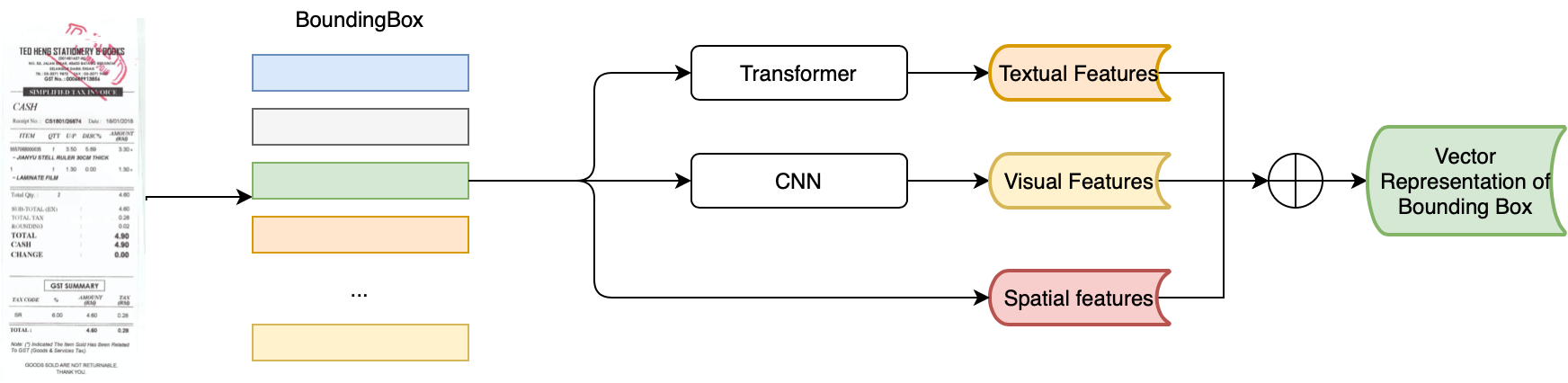}
    \caption{Encoder of architecture}
    \label{encoder}
\end{figure*}
\subsection{Decoder}
\begin{figure}[h]
    \centering
     \includegraphics[height=8cm]{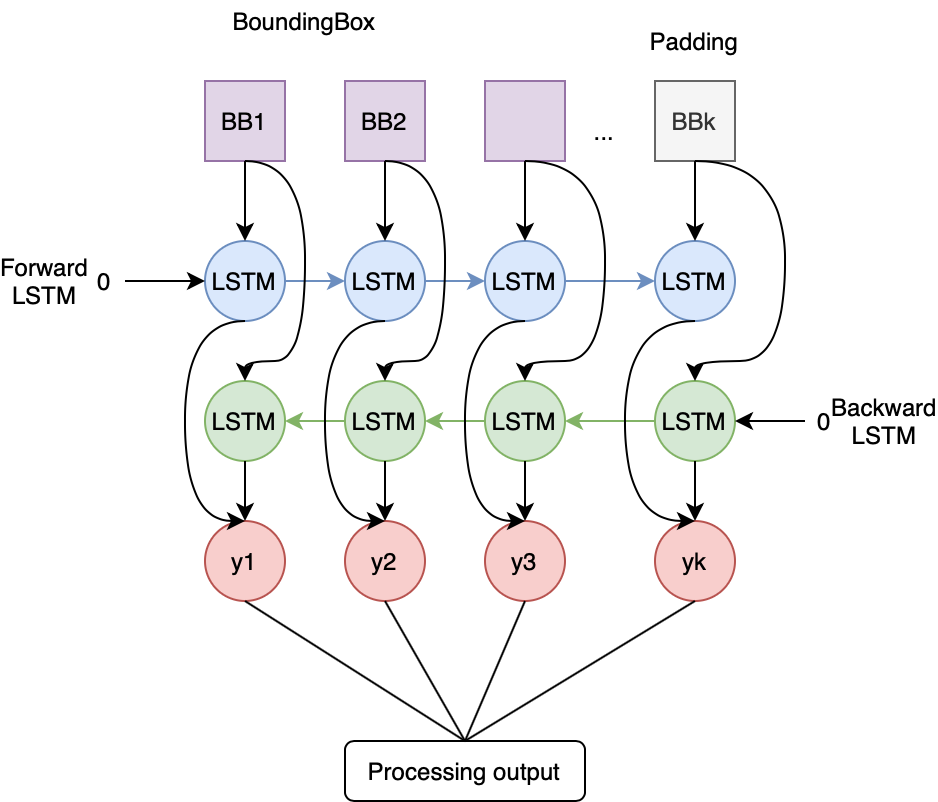}
    \caption{Sequential features in invoices}
    \label{decoder}
\end{figure}
Decoder helps to distinguish between tagging each bounding box. Decoder shown in Figure \ref{decoder}. Decoder consists of BiLSTM and CRF\cite{crf}\cite{bilstm-crf}. 
\[Bouding Box :26/02/1998 18:12:1222\]
Since we are performing bouding box level tagging. Each bouding box may contain some other information, so have to perform post-process for refining tagged bouding box. Bouding box is correctly tagged with correct class but it contain some other information. Using simple refining approach like, template matching we can remove some noise text from tagged bouding box text. For example, bouding box is correctly tagged with Date. However, it contain some redendant information like time, 

\section{Experiment and result}
The complete pipeline for KIPE contain two module. One for encoders, getting all required and helpful features from the bounding box. Second is decoder, for classifying each bounding box into different categories, like total, address, company name, date, invoice number, and etc. 

Encoder: This module extracts textual , visual, and spatial features from the bounding box of invoices. Bounding box i encode with vector Vi. This system encodes each word in a bounding box with some vector using a pre-trained language model, followed by taking the average of each word in the bounding box. Image features ensemble using pre-trained CNN module. In proposed model spatial features can easily calculated as we have all bounding box with text. 

Decoder: Decoder performs tagging each bounding box using BiLSTM. In this way, the proposed model ensembles different features from invoices for extracting key invoice parameters. 
\subsection{Notation}
Each Invoice (I) may contain a variable number of bounding boxes. Bounding box is each block detected with text and enclosed coordinates. Since we have a heterogeneous structure of invoice, we may have a variable number of bounding boxes. For training of BiLSTM in batches, all time series need to have the same length, we need to perform padding till we have some threshold number of bounding boxes. For the proposed model, we set it to the maximum number of bounding boxes from (Extracted text from invoice)ETFI. BBi denotes ith bounding box. I belongs {BB1, BB2, BB3, BB4,.., BBk}. Each bounding box can contain variable number of words. BBi belongs {W1, W2, Wmi}. Textual features of BBi represent by TFi. Visual features represented by VFi. Spatial features represent by SFi. Assembling all features, vector representation of BBi represented by Vi. Yi denotes class of BBi. 

\subsection{Text semantic features}

Taking a simple average, we are giving equal weights to all words, but some of words have more importance to current context semantically and syntactically. Using weighted average, we can represent bounding box with more precisely\cite{sanjiv}

Many model proposed for learning vector space representations of words in capturing fine-grained semantic and syntactic regularities using vector arithmetic.\cite{glove}.
For proposed method, we are using BERT for word embedding. BERT is designed to pre- train deep bidirectional representations from unlabeled text by jointly conditioning on both left and right context in all layers. As a re- sult, the pre-trained BERT model can be fine- tuned with just one additional output layer to create state-of-the-art models for a wide range of tasks\cite{bert}. Word embedding wv1 of word w1 represents using, wv1 = Transformer(W1); Since we have multiple words in the bounding box, textual features of the bounding box are represented by taking the simple average of each word in BB. However in proposed method, we have used transformer BASE BERT model of embedding size 768. 
 During textual information, for embedding name entity, we are using another pretrained model for name entity recognition model, then map that name with \#NAME\#. Continuous set of digits also mapped with \#NUMBER\#. 

\subsection{Visual Features}
Most models use only textual semantic features, but for getting complete insight information from invoices we need to get and use its visual features for invoice parameter extraction. For each bounding box in invoice, visual embedding is denoted as follows,
\[ VFi = CNN(BBi)\] 
where BBi h'xw'x3 belongs to I hxwx3 , denotes a bounding box. Since we have to resize each bounding box according to input from CNN. Visual features extractor module is implemented with ResNet50 for generating visual embedding of bounding box. \cite{pick_29} \cite{resnet}
\subsection{spatial features}

For the proposed model we used some hand picked features.
1. Unit area of BB per total area of invoice document
2. Normalised coordinates of bounding box
3. Unit number of characters per unit area of bounding box.
Our input for the proposed experiment is bounding box from invoices. Since each bounding box is represented with text and its coordinates. Spatial features can easily be calculated, but accuracy and precision of this vector is dependent on how precisely bounding box coordinates are calculated.  

\subsection{Vector representation of bounding box}
Vector representation of each BB Vi calculated by assembling all three features sets that we have calculated. For the proposed experiment, size of output is 768 textual features, 128 image features, and 6 spatial features.

\subsection{Result}

\begin{table}[ht]
\begin{center}
\caption{Performance evaluation of proposed model}
\begin{tabular}{ |c|c| } 
\hline
Method & SROIE Dataset  \\
\hline
Character level BiLSTM & 75.4 \\ 
\hline
Word level BiLSTM & 83.4 \\ 
\hline
Proposed model& 88.2 \\ 
\hline
\end{tabular}
\end{center}
\end{table}

We report that our proposed model out-performed previous benchmark method for extracting key parameters from scanned documents. Since we are assembling visual features and spatial features, this method capable to tag each bounding box with improved accuracy. All over accuracy of complete pipeline is depend the accuracy of detecting bounding box and text in-it. If accuracy of OCR is not good then irrespective of our model accuracy all over accuracy will remain low. Here we have used data-set with bounding box and extracted text in-it. However in real world, it is  difficult to extract bounding box with 100 \% accuracy.

\begin{figure}[h]
    \centering
     \includegraphics[height=2cm]{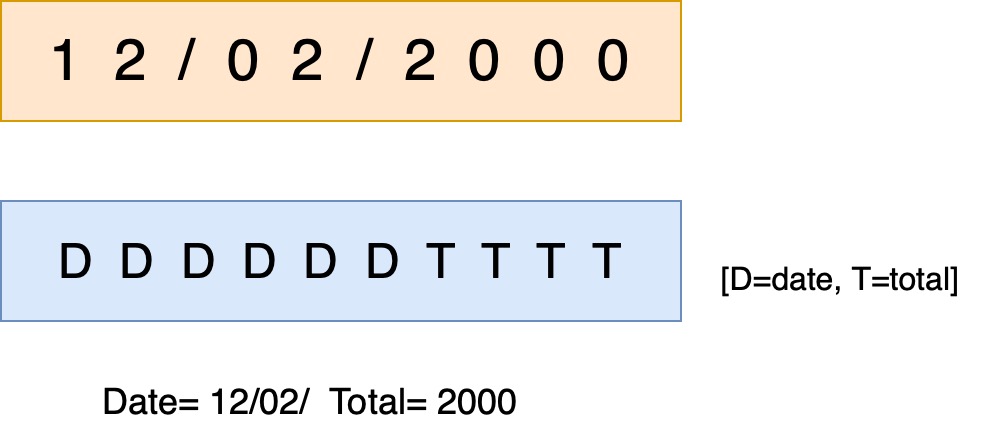}
    \caption{Dis-continuous class tagging in character classification using Bi-LSTM}
    \label{fig:7}
\end{figure}

Word level and character level sequence tagging method for KIPE suffer from problem of wrong tagging or discontinuous tagging. Which may cause distorted output or mis-match in output. In Figure\ref{fig:7}, we can see that final output may have discontinuous class tagging. Using bounding box level class tagging, we can solve this problem. The main contribution of this paper is to focus to exploit visual and some other important features for KIPE and try to implement this method at bounding box level tagging instead of word level or character level classification.
\subsection{Data set}
Our architecture for KIPE can be used for any documents. We have validated this architecture for invoices. Proposed architecture is evaluated on ICDAR 2019 robust reading challenge dataset. This dataset have 1000 scanned receipts. SROIE dataset from the robust reading competition is split into training, validation and test dataset. Out of 1000 scanned receipts, 876 are for training and 347 for validation/test invoices. Where invoices contain pre-extracted bounding boxes with coordinates and text in-it. SROIE datset contain total 1000 invoices, in which 876 are trainable receipts. Each 876 invoices are annoted with invoice parameters. Parameters for this dataset are company name, address, date and total. Each invoice annoted with json file for key invoice parameter. Since testing receipts are not annoted, out of 876 invoices we have removed some invoice with duplicate number and bad annotation and out of remaining we split dataset into 80:20 for training to test purposes.

\section{Discussion}
Abstract information extraction from scanned documents has many advantages in various applications and industries. It helps to streamline data from documents into meaningful form. This paper proposes the model to extract information from scanned documents. Experimental results show performance gain over other methods for Key Information Extraction (KIE). Performance gains mainly power by exploiting unique features from scanned documents, like spatial features, semantic features of text, and visual features of text. This study opens new prospects in field of Document analysis and information extraction.

\bibliographystyle{plain}
\bibliography{references}

\begin{thebibliography}{10}

\bibitem{sanjiv}
Sanjeev Arora, Yingyu Liang, and Tengyu Ma.
\newblock A simple but tough-to-beat baseline for sentence embeddings.
\newblock 2016.

\bibitem{bilstm-crf}
Tao Chen, Ruifeng Xu, Yulan He, and Xuan Wang.
\newblock Improving sentiment analysis via sentence type classification using
  bilstm-crf and cnn.
\newblock {\em Expert Systems with Applications}, 72:221--230, 2017.

\bibitem{bert}
Jacob Devlin, Ming-Wei Chang, Kenton Lee, and Kristina Toutanova.
\newblock Bert: Pre-training of deep bidirectional transformers for language
  understanding.
\newblock {\em arXiv preprint arXiv:1810.04805}, 2018.

\bibitem{pick_29}
Kaiming He, Xiangyu Zhang, Shaoqing Ren, and Jian Sun.
\newblock Deep residual learning for image recognition.
\newblock In {\em Proceedings of the IEEE conference on computer vision and
  pattern recognition}, pages 770--778, 2016.

\bibitem{icdar2019}
Zheng Huang, Kai Chen, Jianhua He, Xiang Bai, Dimosthenis Karatzas, Shijian Lu,
  and CV~Jawahar.
\newblock Icdar2019 competition on scanned receipt ocr and information
  extraction.
\newblock In {\em 2019 International Conference on Document Analysis and
  Recognition (ICDAR)}, pages 1516--1520. IEEE, 2019.

\bibitem{ref14}
Zhiheng Huang, Wei Xu, and Kai Yu.
\newblock Bidirectional lstm-crf models for sequence tagging.
\newblock {\em arXiv preprint arXiv:1508.01991}, 2015.

\bibitem{11pick}
Bo~Jiang, Ziyan Zhang, Doudou Lin, Jin Tang, and Bin Luo.
\newblock Semi-supervised learning with graph learning-convolutional networks.
\newblock In {\em Proceedings of the IEEE Conference on Computer Vision and
  Pattern Recognition}, pages 11313--11320, 2019.

\bibitem{crf}
John Lafferty, Andrew McCallum, and Fernando~CN Pereira.
\newblock Conditional random fields: Probabilistic models for segmenting and
  labeling sequence data.
\newblock 2001.

\bibitem{nameEntity}
Guillaume Lample, Miguel Ballesteros, Sandeep Subramanian, Kazuya Kawakami, and
  Chris Dyer.
\newblock Neural architectures for named entity recognition.
\newblock {\em arXiv preprint arXiv:1603.01360}, 2016.

\bibitem{graphConv}
Xiaojing Liu, Feiyu Gao, Qiong Zhang, and Huasha Zhao.
\newblock Graph convolution for multimodal information extraction from visually
  rich documents.
\newblock {\em arXiv preprint arXiv:1903.11279}, 2019.

\bibitem{gcn}
Devashish Lohani, A~Bela{\"\i}d, and Yolande Bela{\"\i}d.
\newblock An invoice reading system using a graph convolutional network.
\newblock In {\em Asian Conference on Computer Vision}, pages 144--158.
  Springer, 2018.

\bibitem{end2endseq}
Xuezhe Ma and Eduard Hovy.
\newblock End-to-end sequence labeling via bi-directional lstm-cnns-crf.
\newblock {\em arXiv preprint arXiv:1603.01354}, 2016.

\bibitem{glove}
Jeffrey Pennington, Richard Socher, and Christopher~D Manning.
\newblock Glove: Global vectors for word representation.
\newblock In {\em Proceedings of the 2014 conference on empirical methods in
  natural language processing (EMNLP)}, pages 1532--1543, 2014.

\bibitem{graphConv1}
Yujie Qian, Enrico Santus, Zhijing Jin, Jiang Guo, and Regina Barzilay.
\newblock Graphie: A graph-based framework for information extraction.
\newblock {\em arXiv preprint arXiv:1810.13083}, 2018.

\bibitem{tesseract}
Ray Smith.
\newblock An overview of the tesseract ocr engine.
\newblock In {\em Ninth International Conference on Document Analysis and
  Recognition (ICDAR 2007)}, volume~2, pages 629--633. IEEE, 2007.

\bibitem{tesseract1}
Ray Smith.
\newblock Tesseract ocr engine.
\newblock {\em Lecture. Google Code. Google Inc}, 2007.

\bibitem{resnet}
Christian Szegedy, Sergey Ioffe, Vincent Vanhoucke, and Alexander~A Alemi.
\newblock Inception-v4, inception-resnet and the impact of residual connections
  on learning.
\newblock In {\em Thirty-first AAAI conference on artificial intelligence},
  2017.

\bibitem{featureCNN}
Xiao Yang, Ersin Yumer, Paul Asente, Mike Kraley, Daniel Kifer, and
  C~Lee~Giles.
\newblock Learning to extract semantic structure from documents using
  multimodal fully convolutional neural networks.
\newblock In {\em Proceedings of the IEEE Conference on Computer Vision and
  Pattern Recognition}, pages 5315--5324, 2017.

\end{thebibliography}

\end{document}